\documentclass[runningheads]{llncs}
\usepackage{graphicx}
\usepackage{amsmath}
\usepackage{amssymb}
\usepackage{bbding}
\usepackage{CJKutf8}
\usepackage{appendix}

\usepackage{url}   
\usepackage[colorlinks=true,linkcolor=blue,citecolor=blue,urlcolor=blue,anchorcolor=blue,]{hyperref}

\begin{document}
\begin{CJK}{UTF8}{gbsn}
\title{Transformer-Based Person Search with High-Frequency Augmentation and Multi-Wave Mixing}

\titlerunning{High-Frequency Augmentation and Multi-Wave
Mixing}

%
%
\author{Qilin Shu \and Qixian Zhang \and Qi Zhang \and Hongyun Zhang \and Duoqian Miao \and Cairong Zhao} 

\authorrunning{Shu et al.}

\institute{Tongji University}
%
\maketitle              
\begin{abstract}
The person search task aims to locate a target person within a set of scene images. In recent years, transformer-based models in this field have made some progress. However, they still face three primary challenges: 1) the self-attention mechanism tends to suppress high-freq-\\uency components in the features, which severely impacts model performance; 2) the computational cost of transformers is relatively high.
To address these issues, we propose a novel High-frequency Augmentation and Multi-Wave mixing (HAMW) method for person search. HAMW is designed to enhance the discriminative feature extraction capabilities of transformers while reducing computational overhead and improving efficiency. Specifically, we develop a three-stage framework that progressively optimizes both detection and re-identification performance. Our model enhances the perception of high-frequency features by learning from augmented inputs containing additional high-frequency components. Furthermore, we replace the self-attention layers in the transformer with a strategy based on multi-level Haar wavelet fusion to capture multi-scale features. This not only lowers the computational complexity but also alleviates the suppression of high-frequency features and enhances the ability to exploit multi-scale information.
Extensive experiments demonstrate that HAMW achieves state-of-the-art performance on both the CUHK-SYSU and PRW datasets.

\keywords{Person Search  \and Haar Wavelet \and High-Frequency Augmentation.}
\end{abstract}
%
%
%

%
%
%
%

\section{Introduction}

Person search \cite{xiao2017joint,zheng2017person,li2021sequential,zhang2025dynamic,zhang2024learning,zhang2024attentive} aims to locate and identify a specific query person from a set of scene images. This task comprises two subtasks: person detection and person re-identification (ReID) \cite{dou2022human}. Person detection focuses on identifying all persons in the scene and generating bounding-box proposals, while ReID is responsible for identifying the target individual from these proposals.

\begin{figure}[!t]
    \centering
    \includegraphics[width=0.6\linewidth]{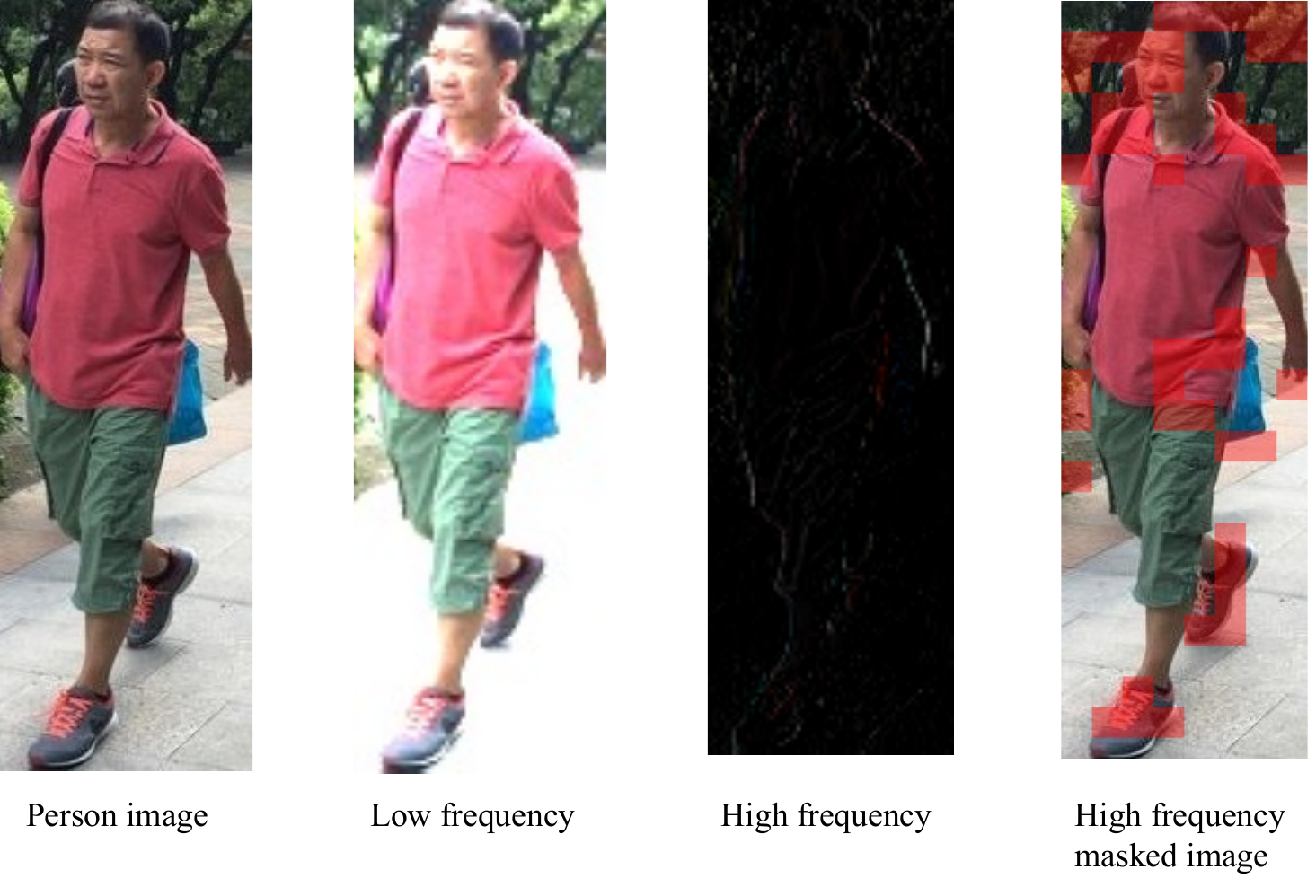}
    \caption{In person search, high-frequency components\textemdash such as image details, edges, and textures\textemdash play a critical role in both pedestrian detection and re-identification.}
    \label{fig1}
\end{figure}

In recent years, transformer-based methods have emerged and shown promising results. Compared to CNN-based approaches, transformer-based models offer unique advantages in terms of discriminative feature learning and robustness to occlusion, pose variation, and scale changes. Nevertheless, the application of transformers faces several challenges, primarily in the following two aspects:

\textbf{Limited ability to capture high-frequency components.} As illustrated in Fig.\ref{fig1}, high-frequency information refers to regions of rapid pixel variation, such as edges and textures that are crucial for preserving fine-grained details. Studies have shown that the self-attention mechanism in transformers tends to attenuate high-frequency signals, leading to performance degradation\cite{zhang2023pha}.

\textbf{High computational complexity.} The quadratic computational cost of self-attention with respect to the input sequence length has prompted the development of techniques such as low-rank and sparse attention. However, in the person search domain, existing methods have yet to effectively address this issue, resulting in transformer-based models that offer limited inference speed advantages compared to their CNN counterparts.

To address the limitations, we propose a novel transformer-based framework for person search, called HAMW. Inspired by COAT \cite{yu2022cascade}, our HAMW adopts a three-stage design that progressively refines detection and ReID performance. The first stage is dedicated to person detection, distinguishing persons from the background. In the second and third stages, ReID and high-frequency augmentation losses are introduced to further enhance identity discrimination. In each stage, we incorporate a branch-augmentation transformer, where the self-attention layers are replaced with multi-wave mixing layers to extract multiscale features, thereby improving discriminability while reducing computational cost.

Moreover, the transformers in the second and third stages introduce high-frequency quantization enhancement (HFQE) and a corresponding high-frequen-\\cy augmentation loss. Tokens augmented with high-frequency components are guided by a proxy-based loss to boost the model's sensitivity to fine-grained details, enhancing its ability to capture and utilize high-frequency information. Experiments on the CUHK-SYSU \cite{xiao2017joint} and PRW \cite{zheng2017person} datasets confirm that our model achieves state-of-the-art performance.

Our main contributions can be summarized as follows.\\
- We propose a novel transformer-based framework for person search that enhances the perception and utilization of high-frequency features while reducing computational complexity.\\
- We replace standard self-attention layers with multi-wave mixing layers, which effectively reduce complexity and improve multi-scale feature representation.\\
- We introduce high-frequency quantization enhancement(HFQE) and a high-frequency augmentation loss, which strengthens the model's ability to represent high-frequency features by bringing closer the tokens of the same identity and their enhanced counterparts while pushing apart those of different identities.\\
- Extensive experiments on two benchmark datasets (CUHK-SYSU and PRW) demonstrate that HAMW achieves 96.0\% mAP and 96.3\% top-1 accuracy on CUHK-SYSU, and 57.5\% mAP and 90.4\% top-1 accuracy on PRW, outperforming previous methods.\\

\section{Method}
\subsection{Overall Architecture}
The overall architecture proposed in this paper is based on COAT\cite{yu2022cascade}, a transfor-\\mer-based multi-stage framework for person search. COAT exhibits excellent performance in handling occlusion and scale variation. Its effectiveness lies in its unique occlusion-aware attention mechanism. During training, this transformer exchanges tokens among all proposals within a batch to simulate real-world occlusions. This strategy enables the model to better adapt to occluded scenarios, learning more robust and discriminative features.

Furthermore, COAT adopts a three-stage cascade design inspired by Cascade R-CNN\cite{cai2018cascade}, allowing the model to refine detection and ReID results progressively from coarse to fine. This approach helps mitigate the conflict between detection and ReID tasks and enhances overall performance. Together, COAT serves as an ideal and promising baseline with ample room for further improvement.

In our model, we retain the cascade structure and token exchange mechanism. To reduce computational complexity and enhance multi-scale feature extraction, we replace the multi-head self-attention layers in the encoder with multi-wave mixing layers. Additionally, to improve the model’s ability to represent high-frequency components in features, we introduce a new auxiliary branch and high-frequency augmentation loss during proposal processing.

\begin{figure}[!t]
    \centering
    \includegraphics[width=1\linewidth]{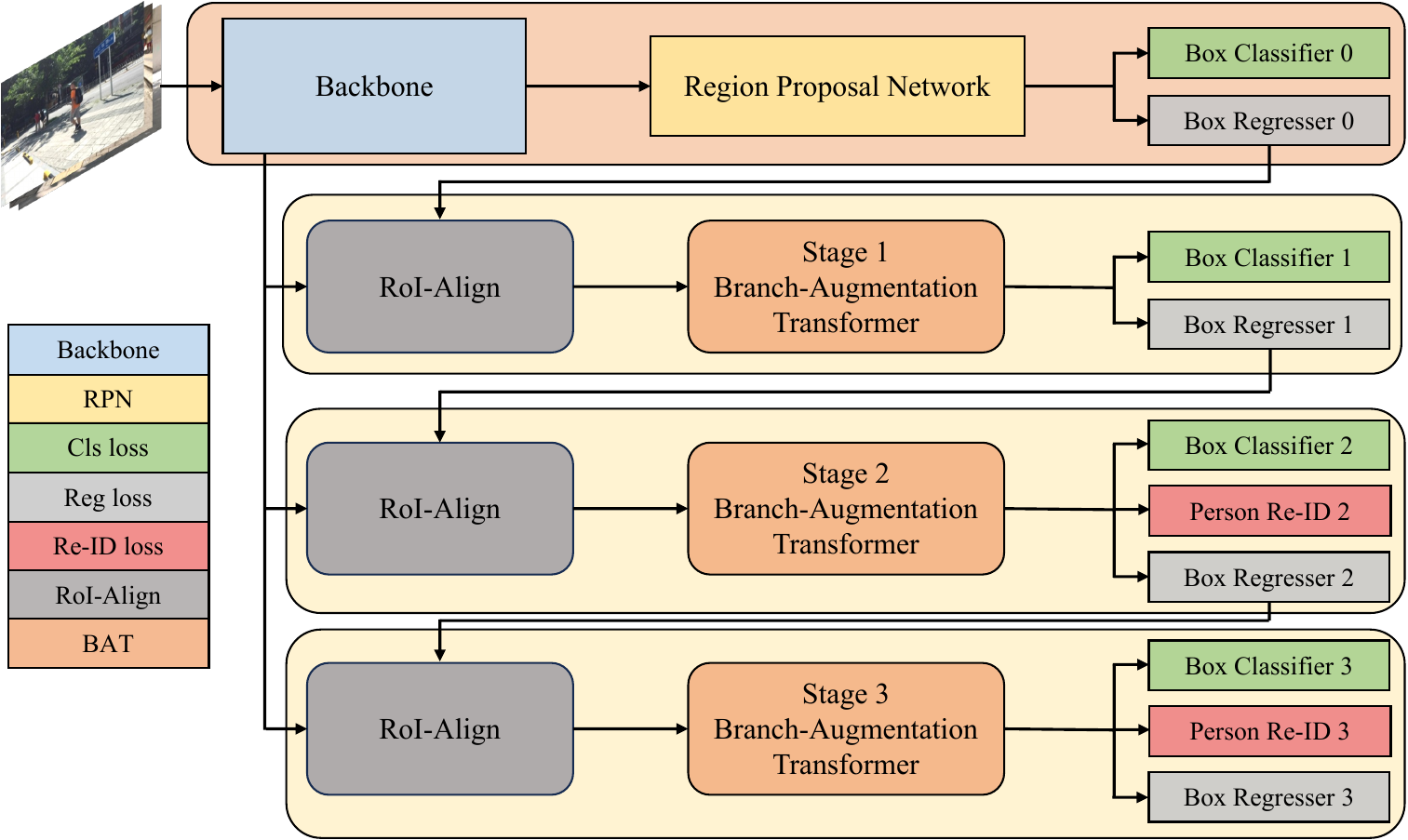}
    \caption{Our overall architecture begins by extracting features from the input image using a backbone network, which are then passed to the Region Proposal Network (RPN). The resulting features and bounding boxes are fed into multiple cascade stages, where RoI-Align pooling is applied. Except for Stage 1, which uses proposals from the RPN, each subsequent stage takes bounding boxes from its preceding stage.}
    \label{fig2}
\end{figure}

As shown in Fig. \ref{fig2}, our framework first extracts features using ResNet-50\cite{he2016deep}, then generates candidate proposals via a Region Proposal Network (RPN)\cite{ren2016faster}. RoI-Align\cite{ren2016faster} is used to pool these proposals of varying sizes into a uniform size. Subsequently, a multi-stage cascade strategy is employed to progressively refine the representations and obtain more precise detection and ReID results. The first stage includes only classification and regression branches, while the second and third stages incorporate an additional ReID branch.

\subsection{Branch-Augmentation Transformer}
\begin{figure}[!t]
    \centering
    \includegraphics[width=1\linewidth]{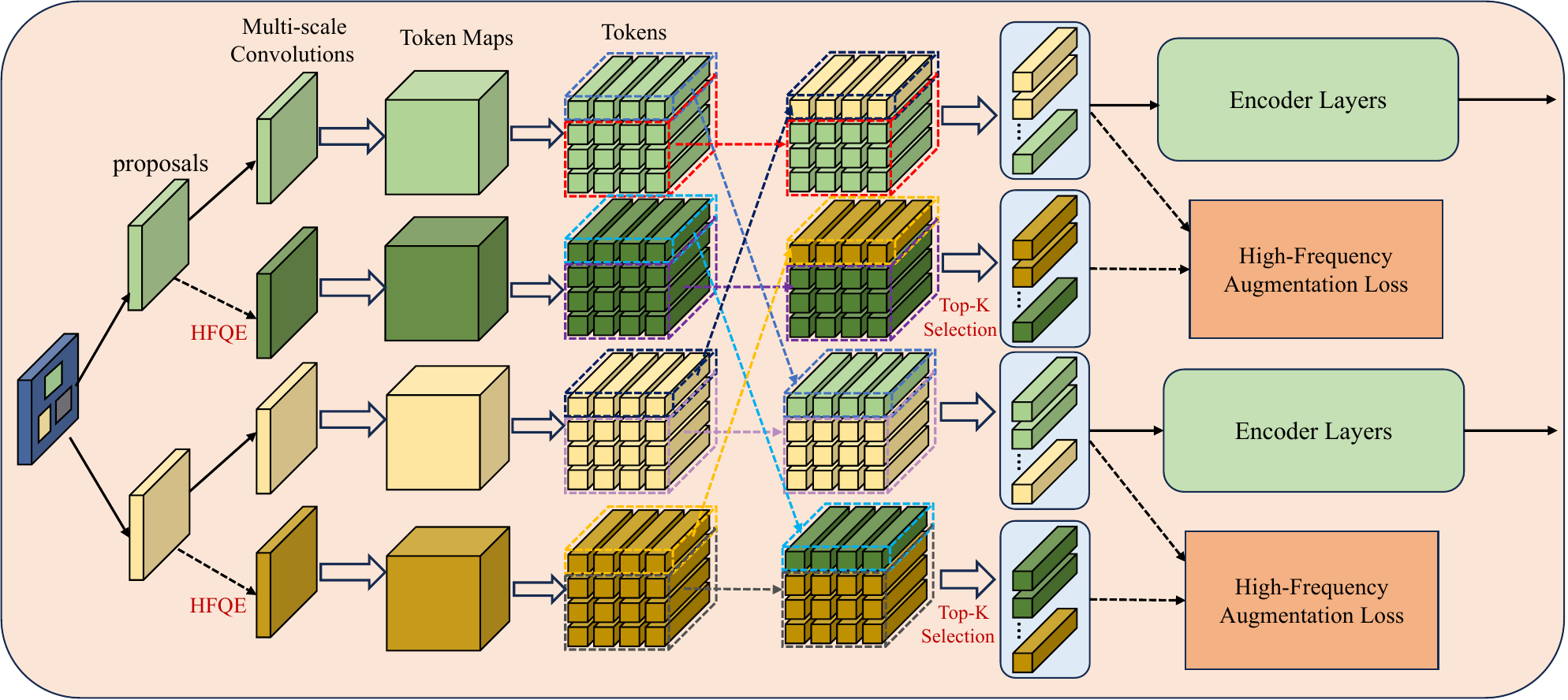}
    \caption{The detailed architecture of the branch-augmentation transformer is illustrated. Note that all operations indicated by black dashed lines are enabled only during the second and third stages of training.}
    \label{fig3}
\end{figure}

\begin{figure}[!t]
    \centering
    \includegraphics[width=0.8\linewidth]{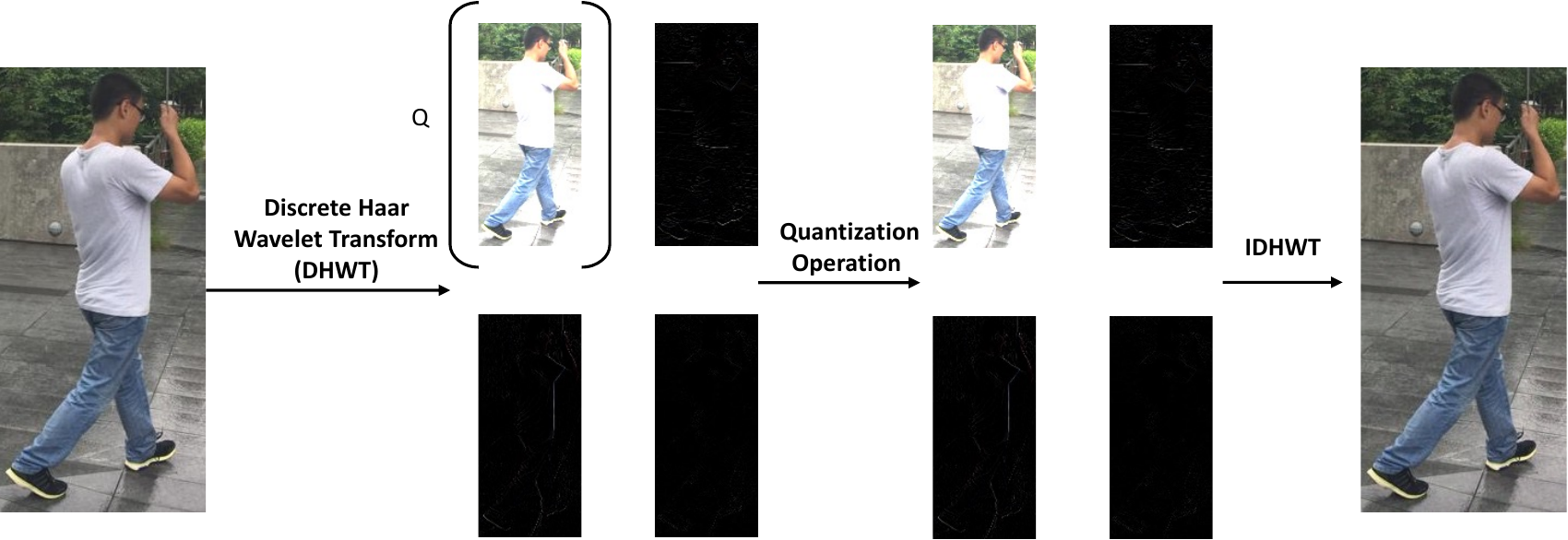}
    \caption{The input of the branch-augmentation transformer is created by applying a 2D DHWT to the features, suppressing the low-frequency (LL) components via quantization, and reconstructing the features, thereby enhancing high-frequency details.}
    \label{fig4}
\end{figure}

One limitation of transformer architecture is that as the number of layers increases, high-frequency components tend to be gradually diluted~\cite{zhang2023pha}. The lack of fine-grained details significantly deteriorates model performance. To address this, we propose a high-frequency augmentation loss in the branch-augmentation transformer to help the model better extract high-frequency features. The architecture of this component is depicted in Fig. \ref{fig3}. For feature maps input into the branch-augmentation transformer, we perform high-frequency quantization enhancement(HFQE). Specifically, for a feature tensor $x\in{\mathcal{R}^{h\text{×}w\text{×}c}}$ we apply 2D Discrete Haar Wavelet Transform (DHWT\cite{mallat2002theory}) to obtain four subbands: $x_{LL},x_{LH},x_{HL},x_{HH}\in{\mathcal{R}^{\frac{h}{2}\text{×}\frac{w}{2}\text{×}c}}$. For the low-frequency subband, we apply quantization as follows:
\begin{equation}
    x_{LL}=\left\lfloor{\frac{x_{L L}+0.5}{q}}\right\rfloor\cdot q
\end{equation}
where $q$ is the quantization interval. We then reconstruct the feature map by applying the inverse 2D DHWT to the four subbands, which effectively suppresses low-frequency components and increases the relative proportion of high-frequency information.Fig.\ref{fig4} briefly illustrates the entire process.

Both the original feature map and the high-frequency enhanced version are sliced along the channel dimension. Each slice is processed by separate convolution layers with distinct parameters, followed by reshaping and reassembling into token maps. Due to these diverse convolutions, each token encodes information from different scales. Next, a partial exchange of tokens occurs within the same batch to simulate occlusion scenarios and enhance robustness. Note that original token maps exchange only with other original token maps, while enhanced token maps exchange only among themselves.

Next, we emphasize the high-frequency augmentation loss. For the enhanced token maps, we again apply 2D DHWT and concatenate the three high-frequency subbands $x_{LH},x_{HL},x_{HH}$:
\begin{equation}
    \mathcal{F}_{h}(x)=\boldsymbol{\operatorname { c o n c a tenate }}\left({x}_{L H}, {x}_{H L},{x}_{H H}\right)
\end{equation}

This operation discards most of the low-frequency information in the enhanced inputs, allowing HAMW to focus on high-frequency signals. We then downsample these new tokens back to their original size and select the top-$K$ tokens with the highest $L_{2}$ norm for computing the high-frequency augmentation loss:
\begin{equation}
    \Omega_{x}:=\{j|j\in\mathrm{top{-}K}(||\mathcal{A}({\mathcal{F}}_{h}(x))||_{2})\} 
\end{equation}
here, $\mathcal{A}$ denotes the downsampling operation. Similarly, we select the top-$K$ tokens from the original token maps based on their $L_{2}$ norm, but without applying downsampling. To compute the loss, we define:
\begin{equation}
    S(x,V,i)=\exp\left(\mathbf{\frac{x_k}{||x_k||_{2}}}^\top\cdot\frac{V_{i,k}^{h}}{\left||V_{i,k}^{h}|\right|_{2}}\right) 
\end{equation}

\begin{figure}[!t]
    \centering
    \includegraphics[width=1\linewidth]{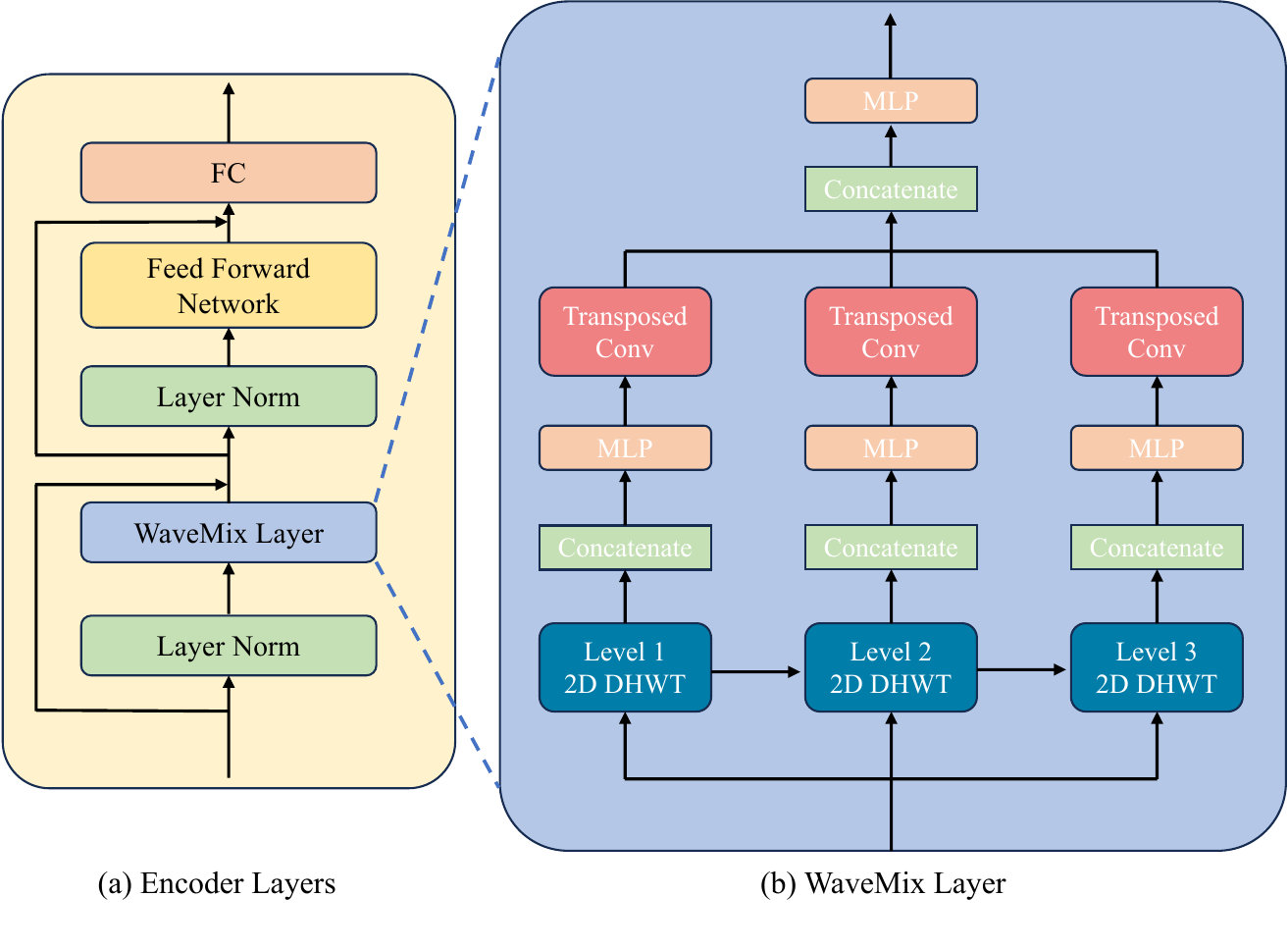}
    \caption{(a) The encoder adopts a standard transformer encoder architecture with WaveMix layers substituted for self-attention, while preserving layer normalization, feed-forward modules, and residual connections.(b) The structure of the WaveMix layer is centered around applying multiple levels of 2D Discrete Haar Wavelet Transforms to the input, followed by concatenation, downsampling, and upsampling operations.}
    \label{fig5}
\end{figure}

Let $x_{k}$ be the token in feature map $x$ ranked $k$-th by $L_{2}$ norm, and $V_{i,k}^{h}$ be the high-frequency enhanced token ranked $k$-th in the table or queue for the $i$-th feature map. Then, the high-frequency augmentation loss is computed as:
\begin{equation}
    \mathcal{L}_{P}=-\sum_{k\in\Omega_{x}}\log\frac{s(x,V,y)}{\sum_{i=1}^{L}\,S(x,V,i)+\sum_{j=1}^{U}S(x,Q,j)}
\end{equation}
where $V$ and $Q$ are the table and queue composed of known and unknown identity samples, respectively, $L$ and $U$ are their lengths, and $y$ denotes the identity index of sample $x$ in $V$. For sample $x$, if a matching identity $V_{y}$ exists in $V$, the loss $\mathcal{L}_{P}$ is calculated and the corresponding entry is updated as:
\begin{equation}
    v_{y}\leftarrow\lambda v_{y}+(1-\lambda)x
\end{equation}
where $\lambda$ is the momentum coefficient. If no matching identity is found in $V$, and the identity of $x$ is known, it is added to $V$; otherwise, it is inserted into the FIFO queue $Q$.All high-frequency quantization enhancement operations are applied only in the second and third stages during training. Afterwards, the original tokens are passed into the encoder layers for further processing. The detailed structure is illustrated in Fig.\ref{fig5}(a).

\subsection{Multi-Wave Mixing Layer}
Existing transformer-based person search models often retain self-attention mec-\\hanisms\cite {yu2022cascade,cao2022pstr}, resulting in high computational complexity. When dealing with long inputs, the computational cost becomes prohibitively high. In contrast, the Discrete Haar Wavelet Transform (DHWT) has linear complexity, maintaining high efficiency regardless of input size. Therefore, we replace the self-attention mechanism with a multi-wave mixing layer. Besides computational efficiency, it also allows for multi-scale feature extraction.

For a tensor $x\in{\mathcal{R}^{h\text{×}w\text{×}c}}$,we first perform a 2D DHWT to obtain four subbands:$x_{LL},x_{LH},x_{HL},x_{HH}\in{\mathcal{R}^{\frac{h}{2}\text{×}\frac{w}{2}\text{×}c}}$,which is the level-1 decomposition. We then apply another 2D DHWT to the low-frequency subband $x_{LL}$ to obtain smaller subbands $y_{LL},y_{LH},y_{HL},y_{HH}\in{\mathcal{R}^{\frac{h}{4}\text{×}\frac{w}{4}\text{×}c}}$
which forms the level-2 decomposition. This process continues until decomposition is no longer possible.

For each level of decomposition, the subbands are concatenated along the channel dimension. Taking level-1 as an example:
\begin{equation}
    \mathcal{M}(x)=\boldsymbol{\operatorname { c o n c a tenate }}\left(x_{LL},{x}_{L H}, {x}_{H L}, {x}_{H H}\right)\in{\mathcal{R}^{\frac{h}{2}\text{×}\frac{w}{2}\text{×}4c}}
\end{equation}

It is then passed through an MLP to restore $M^{'}(x)\in{\mathcal{R}^{\frac{h}{2}\text{×}\frac{w}{2}\text{×}c}}$,followed by a convolutional upsampling to obtain $x^{'}\in{\mathcal{R}^{h\text{×}w\text{×}c}}$.All tensors from different levels are concatenated into $A\in{\mathcal{R}^{h\text{×}w\text{×}nc}}$, where $n$ is the number of decomposition levels. Another MLP is used to project it back to $A'\in{\mathcal{R}^{h\text{×}w\text{×}c}}$,followed by a residual connection. The output is then forwarded to the subsequent layers of the encoder, as illustrated in Fig. \ref{fig5}(b).

\section{Experiments}
In this section, we first introduce the datasets and implementation settings. Then, we compare our HAMW with various approaches. Ablation studies are conducted to evaluate the contribution of each component. Finally, we present more performance analysis to further demonstrate the effectiveness of HAMW.
\subsection{Datasets and Settings}
\subsubsection{Datasets}
We conduct experiments on two widely used person search benchmark datasets: CUHK-SYSU and PRW.

CUHK-SYSU\cite{xiao2017joint} is a large-scale person search benchmark containing 18,184 scene images with 96,143 annotated bounding boxes covering 8,432 identities. It includes images from both street photography and movie scenarios. The street images feature significant variations in viewpoint, lighting, resolution, and occlusion, while movie scenes offer even more diverse and challenging conditions.

PRW\cite{zheng2017person} is a dataset collected with six synchronized cameras on a university campus, focusing on pedestrian detection and re-identification in outdoor scenarios. It includes 11,816 video frames, with 34,304 annotated bounding boxes covering 932 identities.

\subsubsection{Implementation Details}
Our model is implemented using the PyTorch framework and trained on NVIDIA Tesla V100 GPUs. We use the SGD optimizer with a momentum of 0.9 and a weight decay of 5×$10^{-4}$. The model is trained for 20 epochs with a batch size of 3 and an initial learning rate of 0.003. A learning rate warm-up is applied during the first epoch, and the learning rate is decayed after the 15th epoch. During inference, Non-Maximum Suppression (NMS) is used to remove redundant bounding boxes, with NMS thresholds set to 0.4, 0.5, and 0.6 for the three stages, respectively.

\subsection{Comparison with State-of-the-art}
In this subsection, we evaluate HAMW on both benchmark datasets and compare its performance with various methods.


\begin{table*}[!t]
\caption{Comparison with the state-of-the-art methods on CUHK-SYSU and PRW datasets.  The bold entities denote the best performance. 
\label{tab:table1}}
\centering
\renewcommand\arraystretch{1.0}
\tabcolsep=0.4cm
\begin{tabular}{llllll}
\hline
Method             & Backbone          & \multicolumn{2}{l}{CUHKSYSU} & PRW           &               \\ \cline{3-6} 
                   &                   & mAP          & top-1         & mAP           & top-1         \\ \hline
\multicolumn{2}{l}{Two-steps methods}  &              &               &               &               \\
IDE\cite{zheng2017person}                & ResNet50          & -            & -             & 20.5          & 48.3          \\
MGTS\cite{chen2018person}               & VGG16             & 83.0           & 83.7          & 32.6          & 72.1          \\
CLSA\cite{lan2018person}               & ResNet50          & 87.2         & 88.5          & 38.7          & 65.0            \\
RDLR\cite{han2019re}               & ResNet50          & 93.0           & 94.2          & 24.9          & 70.2          \\
IGPN\cite{dong2020instance}               & ResNet50          & 90.3         & 91.4          & 47.2          & 87.0            \\
TCTS\cite{wang2020tcts}               & ResNet50          & 93.9         & 95.1          & 46.8          & 87.5          \\
OR\cite{yao2020joint}                 & ResNet50          & 92.3         & 93.8          & 52.3          & 71.5          \\ \hline
\multicolumn{2}{l}{One-step with CNNs} &              &               &               &               \\
OIM\cite{xiao2017joint}                & ResNet50          & 75.5         & 78.7          & 21.3          & 49.4          \\
RCAA\cite{chang2018rcaa}               & ResNet50          & 79.3         & 81.3          & -             & -             \\
CTXG\cite{yan2019learning}               & ResNet50          & 84.1         & 86.5          & 33.4          & 73.6          \\
NAE\cite{chen2020norm}                & ResNet50          & 91.5         & 92.4          & 43.3          & 80.9          \\
AlignPS+\cite{yan2021anchor}           & ResNet50-DCN      & 94.0           & 94.5          & 46.1          & 82.1          \\
SeqNet\cite{li2021sequential}             & ResNet50          & 94.8         & 95.7          & 47.6          & 87.6          \\
CANR+\cite{zhao2022context}              & ResNet50          & 93.9         & 94.5          & 44.8          & 83.9          \\
SPG\cite{song2023learning}                & ResNet50          & 95.0           & 95.9          & 48.4          & 89.8          \\
SeqNeXt+GFN\cite{jaffe2023gallery}        & ResNet50          & 94.7         & 95.3          & 51.3          & 90.6          \\
DMRNet++\cite{han2022dmrnet++}           & ResNet50          & 94.5         & 95.7          & 52.1          & 87.0            \\ \hline
\multicolumn{3}{l}{One-step with Transformers}        &               &               &               \\
PSTR\cite{cao2022pstr}               & ResNet50          & 93.5         & 95.0            & 49.8          & 87.8          \\
PSTR\cite{cao2022pstr}               & PVTv2-B2          & 95.2         & 96.2          & 56.5          & 89.7          \\
SAT\cite{fiaz2023sat}                & ResNet50          & 95.3         & 96.0            & 55.0            & 89.2          \\
SOLIDER\cite{chen2023beyond}            & Swin-s            & 95.5         & 95.8          & \textbf{59.8} & 86.7          \\
COAT\cite{yu2022cascade}               & ResNet50          & 94.2         & 94.7          & 53.3          & 87.4          \\
ASTD\cite{zhang2024learning}               & ResNet50          & 95.8         & 96.2          & 55.7          & 90.2          \\ \hline
HAMW(Ours)         & ResNet50          & \textbf{96.0}  & \textbf{96.3} & 57.5          & \textbf{90.4} \\ \hline
\end{tabular}
\end{table*}

\subsubsection{Results on CUHK-SYSU}
Table \ref{tab:table1}  presents the results on the CUHK-SYSU dataset. Our HAMW achieves 96\% mAP and 96.3\% top-1 accuracy, outperforming most models. Our approach surpasses the baseline method COAT by +1.8\% mAP and +1.6\% top-1 accuracy. Compared with SOLIDER, which employs semantically controllable self-supervised learning, HAMW also performs better. 

These improvements demonstrate that our model has a stronger capability for perceiving and capturing high-frequency features.

\begin{figure}[!t]
    \centering
    \includegraphics[width=1\linewidth]{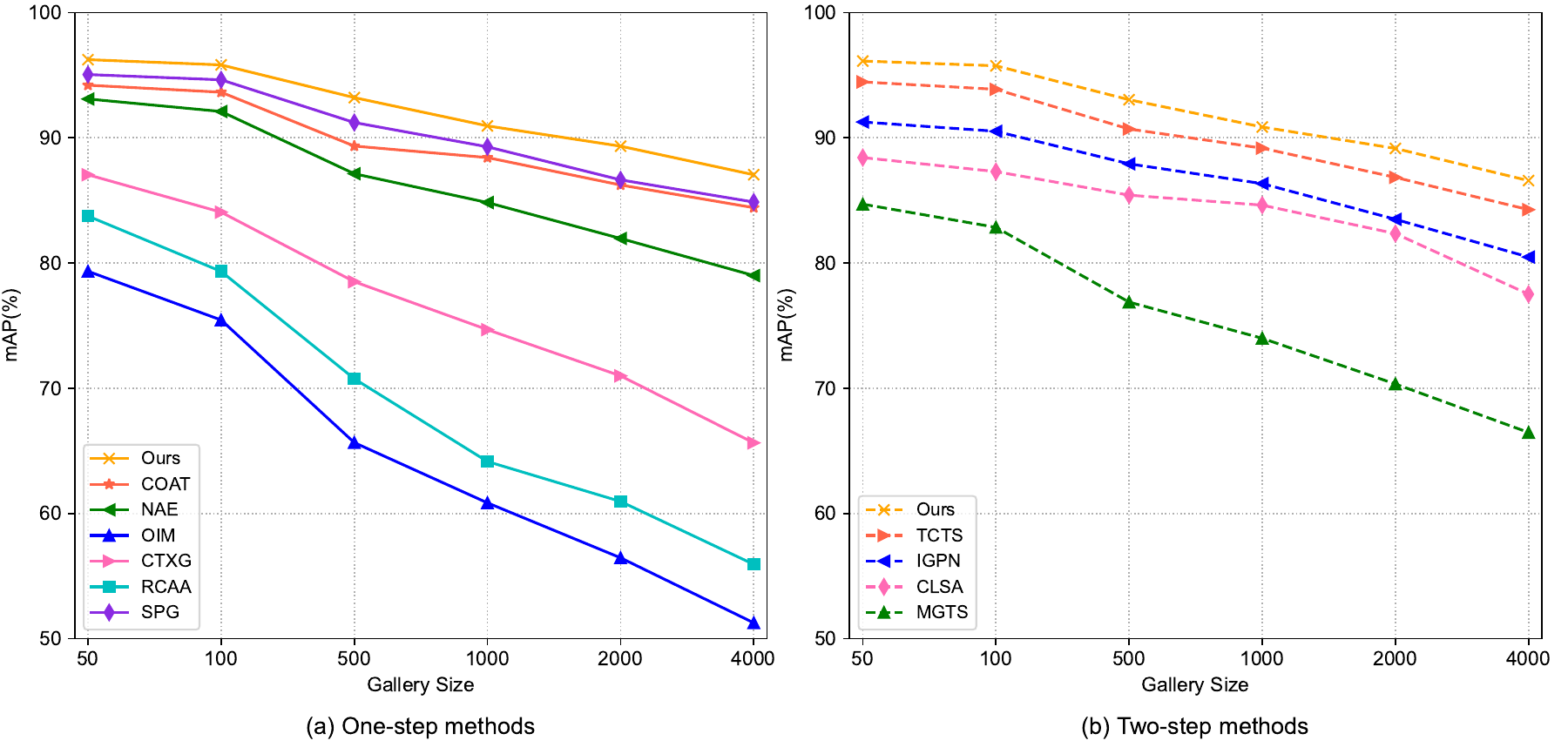}
    \caption{Performance comparison with other methods on the CUHK-SYSU dataset under different gallery sizes: (a) shows the results compared with one-step methods, while (b) presents the results compared with two-step methods.}
    \label{fig6}
\end{figure}

To further validate robustness, we compare HAMW with various approaches under different gallery sizes on the CUHK-SYSU test set. As shown in Fig.\ref{fig6}, performance of all methods drops as the gallery size increases due to the growing number of distractor persons. Nevertheless, HAMW consistently outperforms all baselines, showing strong potential and robustness in large-scale search scenarios.

\subsubsection{Results on PRW}
\begin{figure}[!t]
    \centering
    \includegraphics[width=1\linewidth]{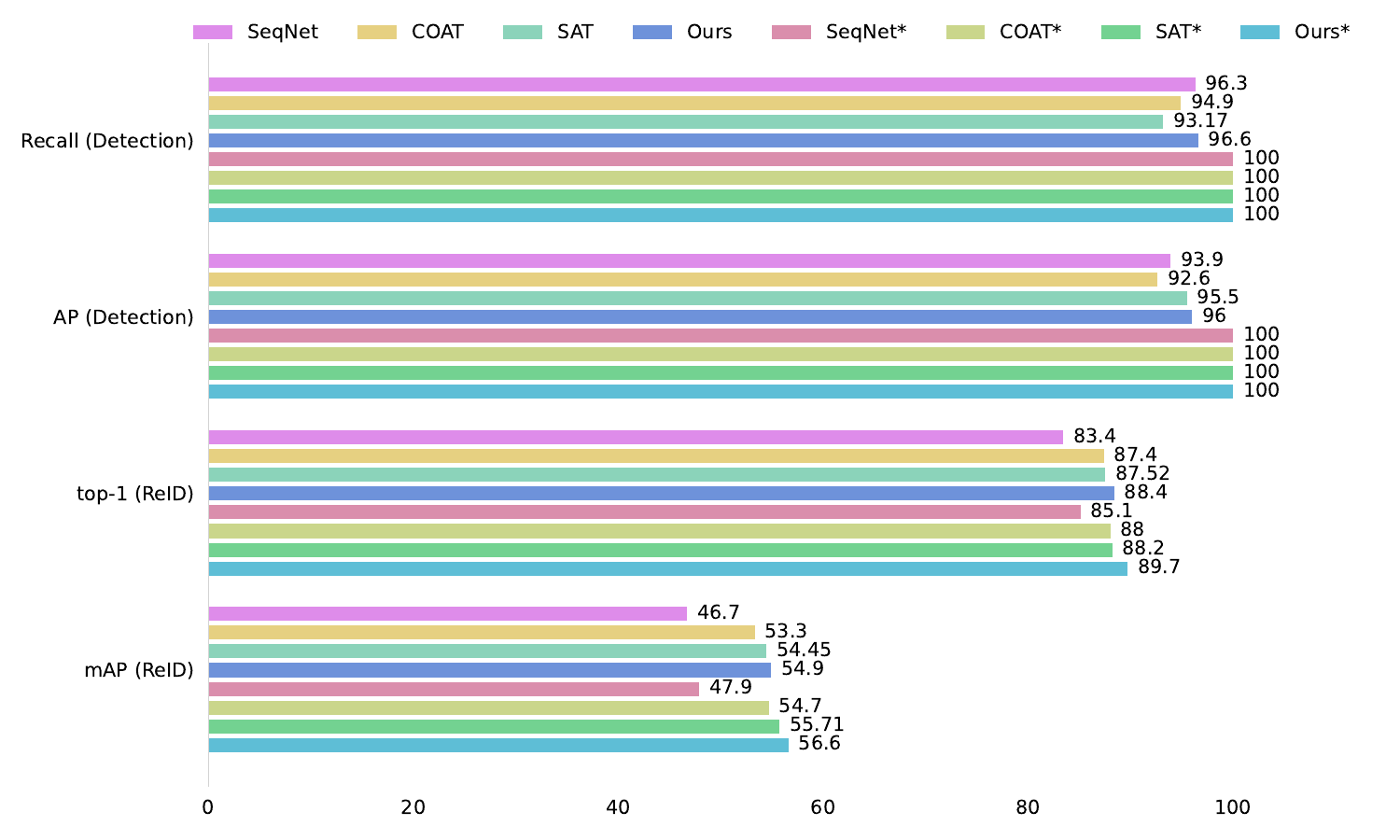}
    \caption{Comparison of person search and detection scores on the PRW dataset with and without using ground-truth annotations. * indicates the use of ground-truth.}
    \label{fig7}
\end{figure}

Table \ref{tab:table1}  reports the performance on the PRW dataset, which is more challenging than CUHK-SYSU due to fewer training samples, a larger gallery, and many visually similar identities.
Our HAMW achieves 57.5\% mAP and 90.4\% top-1 accuracy. Compared to CNN-based methods such as AlignPS and SeqNet, our HAMW outperforms them significantly. It also improves upon COAT by +4.2\% mAP and 3\% top-1 accuracy. In Fig. \ref{fig7}, we further compare HAMW against SeqNet, COAT, and SAT with and without using ground-truth boxes on PRW. The results indicate that HAMW not only achieves superior performance on person search but also benefits from ground-truth information, suggesting its strong capability in handling the re-identification sub-task.

\subsection{Ablation Study}
In this section, we conduct detailed ablation studies to evaluate the design choices of our model. We assess the effectiveness of each component and analyze the contribution of different stages. Additionally, we show that the proposed multi-wave mixing layers effectively reduce inference time.

\subsubsection{Contribution of High-Frequency Augmentation Mechanism}
To investigate the joint contribution of high-frequency quantization enhancement, low-frequency discarding (i.e., top-K selection), and high-frequency augmentation loss, we analyze the effect of introducing them at different stages on the PRW dataset. As shown in Table \ref{tab:table3}, introducing the high-frequency enhancement loss only at stage 2 leads to a 1.3\% improvement in mAP and 1.5\% in top-1 accuracy. Further introducing it at stage 3 yields an additional 0.6\% mAP gain and 0.7\% improvement in top-1. These results demonstrate the effectiveness of enhancing high-frequency components via the proposed loss. The continued performance gains when applied in multiple stages further confirm the cascaded synergistic effect of the high-frequency augmentation mechanism.

As for the individual contributions of the three components, we note that without the high-frequency enhancement loss, the other two (high-frequency quantization enhancement and low-frequency discarding) have no effect. Therefore, we only evaluate the individual impact of the latter two when the loss is present. As shown in Table \ref{tab:table4}, incorporating either high-frequency quantization enhancement or low-frequency discarding independently improves the model's performance. When both are applied together, further improvements are observed. This indicates that each component is effective, and their combination leads to additional performance gains.

\begin{table*}[!t]
\caption{An ablation study on the PRW dataset to analyze the stage-wise impact of high-frequency quantization enhancement, low-frequency discarding, and high-frequency augmentation loss.
\label{tab:table3}}
\centering
\renewcommand\arraystretch{1.0}
\tabcolsep=0.4cm
\begin{tabular}{lllll}
\hline
Index & Stage 2 & Stage 3 & PRW           &               \\ \cline{4-5} 
      &         &         & mAP           & top-1         \\ \hline
1     &         &         & 55.6          & 88.1          \\
2     & \Checkmark      &         & 56.9          & 89.6          \\
3     & \Checkmark      & \Checkmark       & \textbf{57.5} & \textbf{90.3} \\ \hline
\end{tabular}
\end{table*}

\begin{table*}[!t]
\caption{Ablation study of high-frequency quantization enhancement and low-frequency discarding on the PRW dataset, both applied in the second and third stages.
\label{tab:table4}}
\centering
\renewcommand\arraystretch{1.0}
\tabcolsep=0.4cm
\begin{tabular}{lllll}
\hline
Index & HQE & LD & PRW           &               \\ \cline{4-5} 
      &     &    & mAP           & top-1         \\ \hline
1     &     &    & 55.6          & 88.1          \\
2     & \Checkmark  &    & 56.8          & 89.4            \\
3     &     & \Checkmark  & 56.4          & 89.2          \\
4     & \Checkmark   & \Checkmark & \textbf{57.5} & \textbf{90.3} \\ \hline
\end{tabular}
\end{table*}

\subsubsection{Contribution of Multi-Wave Mixing Layer}
To evaluate the effectiveness of the multi-wave mixing layer in terms of both performance and efficiency, we replace them with corresponding modules from various architectures, including CvT\cite{wu2021cvt}, Swin Transformer\cite{liu2021swin}, PVT\cite{wang2021pyramid}, MLP-Mixer\cite{tolstikhin2021mlp}, and GFNet\cite{rao2021global}, and compare the results on the PRW dataset. The results show that our proposed multi-wave mixing layers consistently outperform all these alternatives in both accuracy and inference speed, demonstrating their superior effectiveness. The detailed performance of different self-attention alternatives on the PRW dataset is presented in Table \ref{tab:table5}. For a more detailed comparison of the inference speed of our method, please refer to the Appendix \ref{qualitative analisis}.

\begin{table*}[!t]
\caption{Comparison of inference speed and accuracy on the PRW dataset among different self-attention alternatives.
\label{tab:table5}}
\centering
\renewcommand\arraystretch{1.0}
\tabcolsep=0.4cm
\begin{tabular}{llll}
\hline
Layer               & Time(ms)    & PRW           &               \\ \cline{3-4} 
                    &             & mAP           & top-1         \\ \hline
CvT\cite{wu2021cvt}                 & 78          & 56.2          & 87.4          \\
Swin Transformer\cite{liu2021swin}    & 75          & 57.0            & 89.7          \\
PVT\cite{wang2021pyramid}                 & 83          & 55.8          & 88.9          \\
MLP-Mixer\cite{tolstikhin2021mlp}           & 62          & 56.1          & 87.9          \\
GFNet\cite{rao2021global}               & 55          & 56.6          & 88.2          \\ \hline
Multi-Wave Mixing Layer & \textbf{31} & \textbf{57.5} & \textbf{90.3} \\ \hline
\end{tabular}
\end{table*}

\section{Conclusion}
We propose a transformer-based end-to-end person search model that enhances the perception and extraction of high-frequency and multi-scale information while reducing computational cost and improving inference speed. Specifically, we introduce a high-frequency augmentation loss, enabling the model to learn stronger sensitivity to high-frequency components from specially augmented inputs. Furthermore, we replace the self-attention layers in the transformer encoder with multi-wave mixing layers, which not only preserve the global receptive field characteristic of the transformer but also significantly reduce computational complexity and improve the extraction of multi-scale information.
Extensive experiments on CUHK-SYSU and PRW datasets demonstrate that our HAMW achieves state-of-the-art performance, validating its effectiveness and practicality for real-world person search applications.
 

\bibliographystyle{splncs04_unsort}
\bibliography{reference}






\newpage
\appendix
\setcounter{section}{0}
\renewcommand\thesection{\Alph{section}}

\section{Qualitative Analysis}
\label{qualitative analisis}
\begin{figure}[!ht]
    \centering
    \includegraphics[width=1\linewidth]{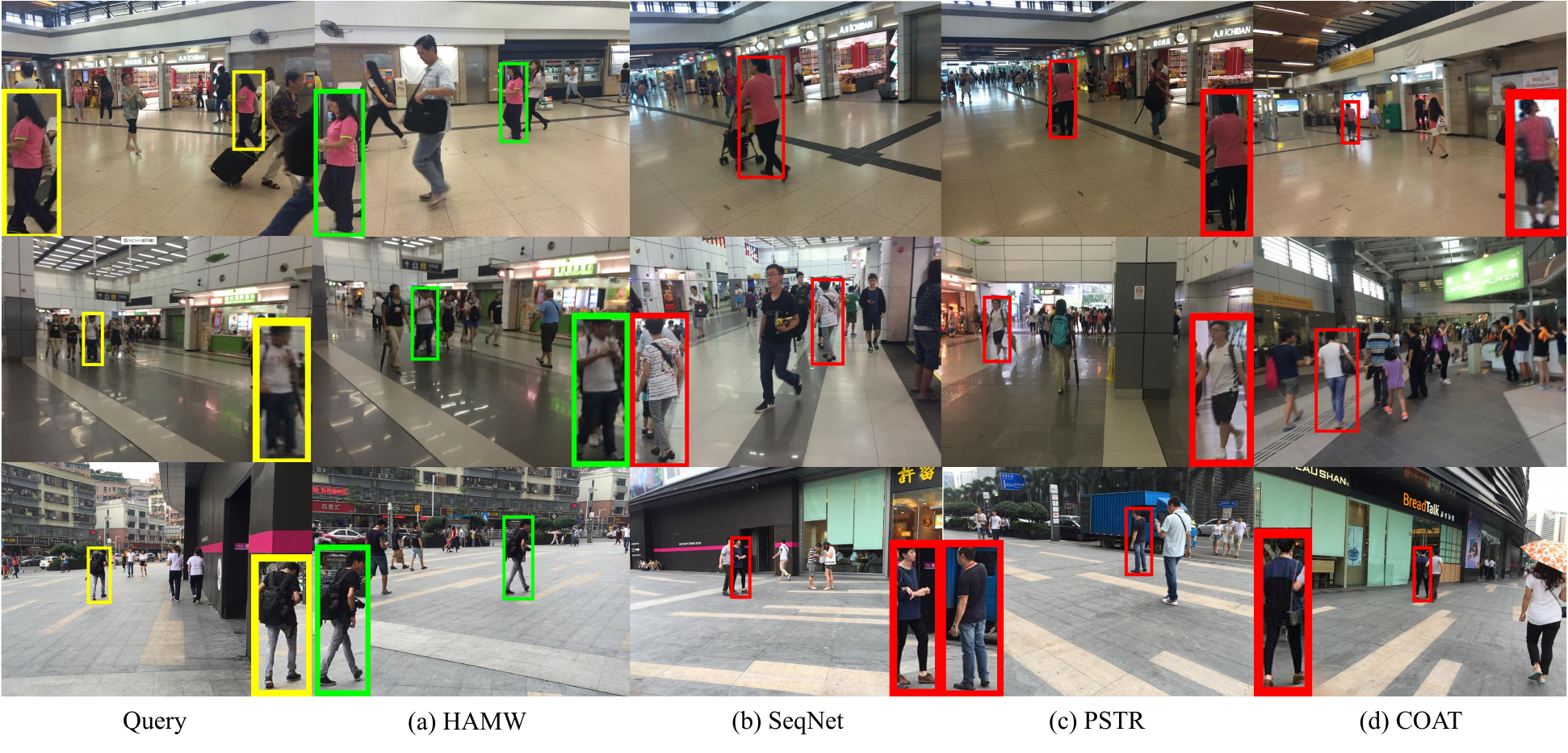}
    \caption{Top-1 person search qualitative results of THAWM, SeqNet, PSTR, and COAT on the CUHK-SYSU dataset. Yellow, green, and red boxes indicate the query, correct matches, and incorrect matches, respectively.}
    \label{fig8}
\end{figure}

\begin{figure}[!ht]
    \centering
    \includegraphics[width=1\linewidth]{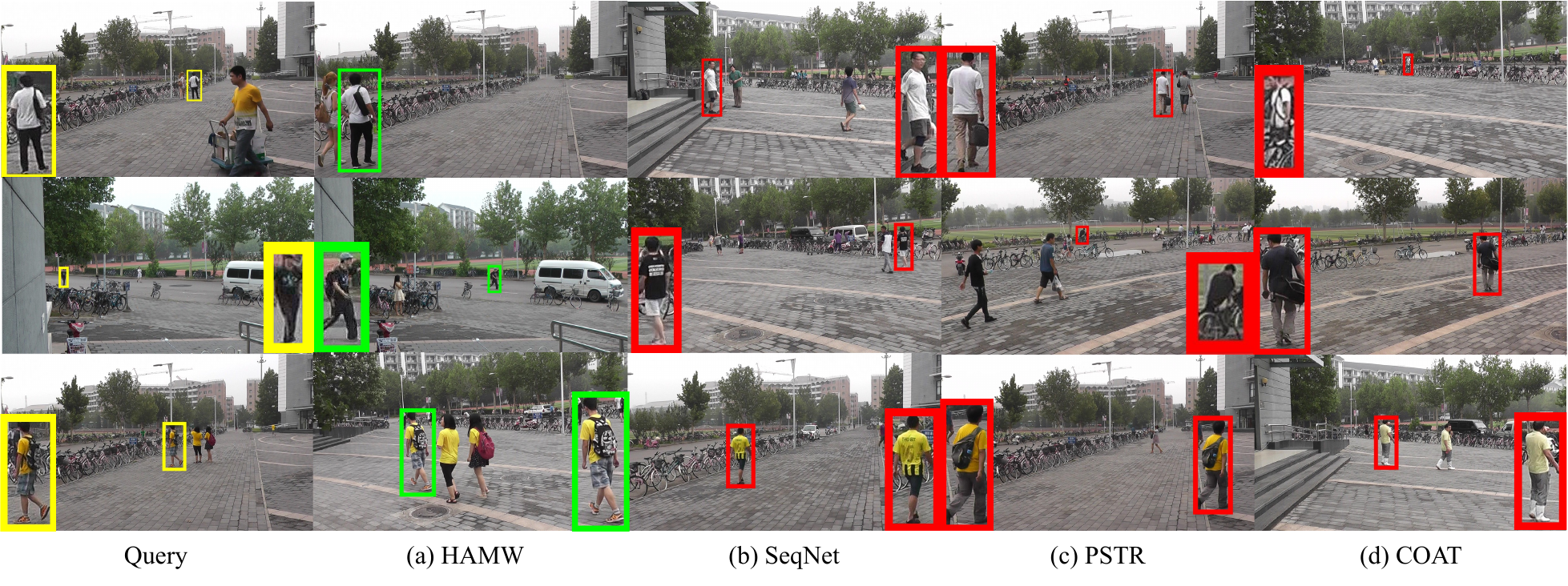}
    \caption{Top-1 person search qualitative results of THAWM, SeqNet, PSTR, and COAT on the PRW dataset. Yellow, green, and red boxes indicate the query, correct matches, and incorrect matches, respectively.}
    \label{fig9}
\end{figure}

We conduct qualitative analysis on the PRW dataset and compare HAMW with SeqNet, PSTR, and COAT. As shown in Fig.\ref{fig8} and Fig.\ref{fig9}, other methods are prone to misidentifications when persons have similar colors or are too small, suggesting that they fail to adequately capture high-frequency details and thus struggle to extract discriminative features. In contrast, our model HAMW, trained with high-frequency enhanced features, can more accurately detect subtle cues on persons and avoid such errors.

\section{Hyperparameter Analysis}
\begin{figure}
    \centering
    \includegraphics[width=1.0\linewidth]{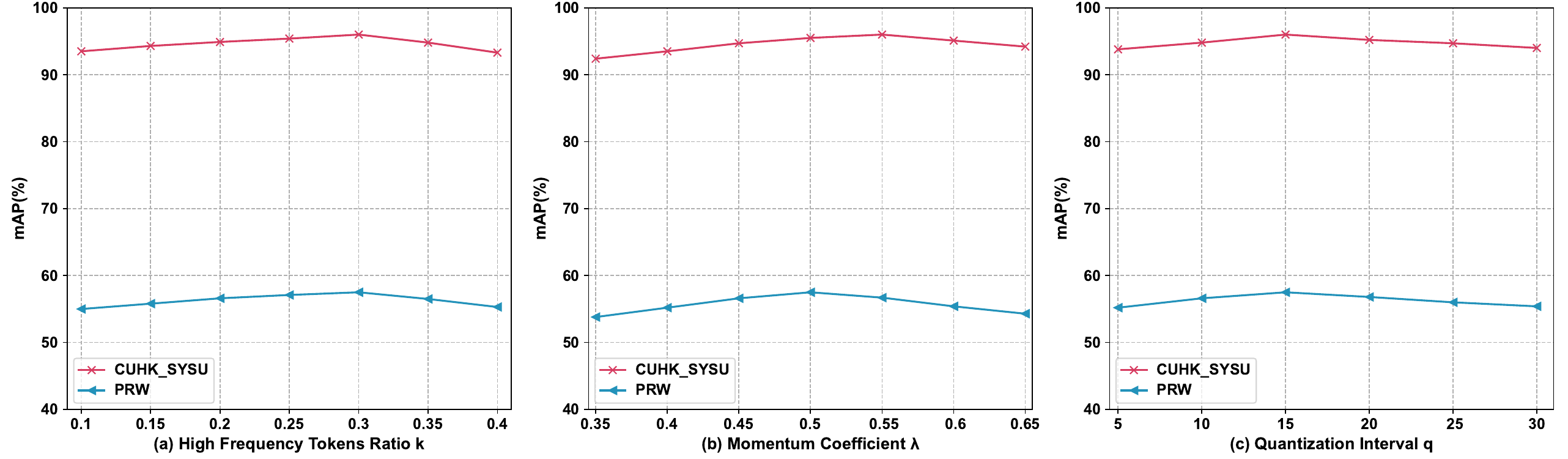}
    \caption{Model performance under different hyperparameter settings on CUHK-SYSU and PRW datasets.}
    \label{fig10}
\end{figure}

Ratio of High-Frequency Tokens $k$: The parameter $k$ controls the ratio of tokens selected for computing the high-frequency augmentation loss. The number of tokens in the feature map multiplied by $k$ gives the value of $K$ mentioned earlier. It reflects the assumed ratio of high-frequency information within the entire feature map. As shown in Fig.\ref{fig10}(a), the model achieves the best performance when $k$=0.3. A larger $k$ may mistakenly include some low-frequency components, which could degrade the performance.\\
Momentum Coefficient $\lambda$: The coefficient $\lambda$ is used for updating the Table $V$ during the computation of the high-frequency augmentation loss. It balances the importance between new and old information: smaller $\lambda$ values make the model more sensitive and adaptive, while larger values lead to more stable and noise-resistant behavior. As shown in Fig.\ref{fig10}(b), both $\lambda$=0.5 and $\lambda$=0.55 yield comparable and optimal results.\\
Quantization Interval $q$: The parameter $q$ controls the degree of quantization applied to the low-frequency subbands. A larger $q$ leads to more aggressive suppression of low-frequency components. While removing low-frequency information can be beneficial, it is also crucial to retain meaningful low-frequency cues. As shown in Fig.\ref{fig10}(c) indicate that the model achieves the highest accuracy when $q$=15. Larger values of $q$ may start to harm the model's performance.\\

\subsubsection*{Efficiency Comparison}
To demonstrate the efficiency of HAMW, we report the inference time (in milliseconds) on a Tesla V100 GPU. For fair comparison, all input images are resized to 900 × 1500. As shown in Table \ref{tab:table2}, HAMW achieves the fastest inference speed. This improvement is mainly attributed to replacing the self-attention mechanism with the proposed multi-wave mixing layers.

\begin{table*}[!t]
\caption{The inference speed of different methods on the PRW dataset is summarized to compare their computational efficiency.
\label{tab:table2}}
\centering
\renewcommand\arraystretch{1.0}
\tabcolsep=0.8cm
\begin{tabular}{llll}
\hline
Method     & Backbone & GPU    & Time(ms)    \\ \hline
OIM\cite{xiao2017joint}        & ResNet50 & V100   & 118         \\
NAE\cite{chen2020norm}        & ResNet50 & V100   & 83          \\
MGTS\cite{chen2018person}       & ResNet50 & K800   & 1269        \\
AlignPS+\cite{yan2021anchor}   & ResNet50 & V100   & 67          \\
SeqNet\cite{li2021sequential}     & ResNet50 & V100   & 86          \\
PSTR\cite{cao2022pstr}       & ResNet50 & V100   & 56          \\
COAT\cite{yu2022cascade}       & ResNet50 & V100   & 90          \\
ASTD\cite{zhang2024learning}       & ResNet50 & V100   & 96          \\
SAT\cite{fiaz2023sat}        & ResNet50 & V100   & 105         \\ \hline
HAMW(OURS) & ResNet50 & V100   & \textbf{31} \\ \hline
\end{tabular}
\end{table*}

\section{Loss Function}
During training, our model is supervised by a combination of three loss functions: detection loss $\mathcal{L}_{det}$, re-identification losses $\mathcal{L}_{OIM}$ and $\mathcal{L}_{ID}$ , and the high-frequency augmentation loss $\mathcal{L}_{P}$.

The detection loss $\mathcal{L}_{det}$ includes a cross-entropy loss for distinguishing persons from the background, and a Smooth-L1 loss for refining bounding box coordinates.
The OIM loss $\mathcal{L}_{OIM}$ maintains a lookup table (LUT) for labeled identities and a circular queue (CQ) for unlabeled ones, optimizing the embedding via cosine similarity. The identity loss $\mathcal{L}_{ID}$, also a cross-entropy loss, provides an additional supervision signal by predicting the identity class of persons.
As described earlier, the high-frequency augmentation loss $\mathcal{L}_{P}$ enhances the model’s sensitivity to high-frequency features by leveraging enhanced inputs as auxiliary signals and computing a proxy-based loss on similarity to known or unknown identities.

Among the three losses, the detection loss is applied in all three stages, while the ReID and high-frequency augmentation losses are applied only in the second and third stages. The total loss function is defined as:
\begin{equation}
    \mathcal{L}=\sum_{t=1}^{T}\mathcal{L}_{\mathrm{det}}^{t}+\mathcal{I}(t > 1)(\lambda_{\mathrm{OlM}}\mathcal{L}_{\mathrm{OlM}}^{t}+\lambda_{\mathrm{ID}}\mathcal{L}_{\mathrm{ID}}^{t}+\lambda_{\mathrm{P}}\mathcal{L}_{\mathrm{P}}^{t})
    \label{eq8}
\end{equation}
where $\lambda_{OIM}$, $\lambda_{ID}$ and $\lambda_{P}$ are weighting coefficients to balance the contributions of each loss.

\section{Related Work}
\subsection{Person Search}
Person search methods can be broadly categorized into two main paradigms based on their architectural design:

\textbf{Two-step methods.}These approaches divide the person search task into two independent subtasks: pedestrian detection and person re-identification (Re-ID), each handled by separately trained models. Zheng et al. were the first to combine different pedestrian detectors with various ReID models, pioneering this research direction. Chen et al. proposed the Mask-Guided Two-Stream CNN Model (MGTS) and highlighted the issue of target conflict. Lan et al. introduced Cross-Level Semantic Alignment (CLSA) in the ReID stage to address the multi-scale feature alignment problem. Dong et al. proposed the Instance-guided Proposal Network (IGPN), which incorporates query information into the detection module.

\textbf{One-step methods.}These methods integrate detection and ReID into a single end-to-end trainable framework. They generally offer lower computational cost and higher efficiency. Xiao et al. first introduced an end-to-end model based on Faster R-CNN. Dong et al. designed a bidirectional interaction network to suppress redundant contextual information. Munjal et al. were the first to incorporate query information, proposing the Query-Guided End-to-End Person Search (QEEPS) framework. Yan et al. developed AlignPS, an anchor-free person search model that reduces computational overhead while addressing the challenges of multi-scale and region misalignment in feature reconstruction. Li and Miao introduced SeqNet, a sequential framework using two Faster R-CNNs to handle detection and ReID in order. Jaffe et al. further extended this with SeqNeXt by replacing Faster R-CNN with ConvNeXt and reducing similar background scenarios to lower the search space and computation. Zhang et al. proposed AMPN and ASTD frameworks to tackle scale variation, occlusion, and pose changes, achieving state-of-the-art performance. Additionally, Zhang et al. introduced PS-DFSI, which fuses frequency and spatial information to enhance robustness in person search.
\subsection{Transformers in Person Search}
Since Vision Transformers (ViT) demonstrated their potential in computer vision tasks, many subfields—including person search and person ReID—have explored their use. He et al. were the first to adopt a pure transformer architecture for ReID, using shuffled and rearranged patch embeddings to generate more discriminative and robust features. Wang et al. proposed NFormer, which enhances robustness and discrimination by explicitly modeling relationships between all input images. Luo et al. combined spatial transformer networks (STN) with a ReID module in STNReID to address the misalignment between partial and holistic images during alignment and feature extraction. Li et al. introduced the PAT model, the first to apply transformer architecture to occluded person ReID, also leveraging weakly supervised learning. Zhang et al. proposed the HAT framework, combining CNNs for local feature extraction and transformers for global features, aiming to overcome the limited receptive field of CNNs and enhance global feature discrimination.

\subsection{Improvements to Self-Attention}
While self-attention is the cornerstone of transformer architectures, it suffers from high computational complexity and limited ability to capture local textures and multi-scale information. Numerous studies have proposed modifications to mitigate these issues. d’Ascoli et al. introduced Gated Positional Self-Attention (GPSA), which initializes as a convolutional layer and gradually learns attention through gating. Dai et al. proposed CoAtNet, combining DW-Conv with multi-head attention, arranged in alternating layers. Liu et al. developed window-based self-attention by partitioning the image into non-overlapping local windows and cyclically shifting them to enable cross-window interaction, resulting in linear complexity with respect to image size. Wang et al. constructed a Pyramid Vision Transformer (PVT), which progressively reduces token length while employing Spatial-Reduction Attention (SRA) to cut computation, making PVT suitable for high-resolution dense prediction with preserved global receptive fields. Zhu et al. introduced learnable sparse attention in DETR, where each query attends to a small set of sampled points, accelerating convergence and improving small object detection.

Tolstikhin et al. proposed a pure MLP-based vision model (MLP-Mixer), which alternates between token-mixing and channel-mixing MLPs to aggregate spatial and channel-wise information, achieving comparable performance to transformers under large-scale pretraining. Rao et al. introduced GFNet, replacing self-attention with global filtering via 2D FFT and learned frequency-domain filters, followed by inverse FFT to return to the spatial domain—showing promise as a transformer alternative on ImageNet and downstream tasks. Guibas et al. proposed AFNO, drawing on Fourier Neural Operators to model token interactions in the frequency domain via continuous convolutions, achieving near-linear complexity.

In contrast, our HAMW retains the cascaded architecture of COAT but introduces High-Frequency Quantization Enhancement (HFQE) to augment the input with additional high-frequency content, thereby improving the model's ability to perceive fine-grained features. Furthermore, we replace the traditional self-attention layers in transformers with multi-wave mixing layers, which preserve the global receptive field while significantly reducing computational cost. The hierarchical structure also provides additional advantages in capturing multi-scale information effectively.

\end{CJK}
\end{document}